\documentclass[10pt,twocolumn,letterpaper]{article}

\usepackage{iccv}
\usepackage{times}
\usepackage{epsfig}
\usepackage{graphicx}
\usepackage{amsmath}
\usepackage{amssymb}

\usepackage{balance}
\usepackage{flushend}

\usepackage{color}
\usepackage[utf8]{inputenc}
\usepackage{multirow}
\usepackage{array}
\usepackage{adjustbox}
\usepackage{bbding}
\usepackage{makecell}

\usepackage{pifont}

\usepackage{amsmath}
\usepackage{amssymb}
\usepackage{array}
\usepackage{adjustbox}
\usepackage{booktabs}
\usepackage{multirow}

\usepackage{color}
\usepackage{balance}
\usepackage{flushend}

\usepackage{CJKutf8}

\usepackage{algorithm}
\usepackage{algorithmic}
\usepackage{multirow}

\usepackage[pagebackref=true,breaklinks=true,letterpaper=true,colorlinks,bookmarks=false]{hyperref}

\iccvfinalcopy 

\def\iccvPaperID{2505} 

\ificcvfinal\fi

\begin{document}
\begin{CJK}{UTF8}{gbsn}

\pagestyle{plain}

\title{Mode-locking Theory for Long-Range Interaction in Artificial Neural Networks}

\author{Xiuxiu Bai\thanks{Corresponding author}  , Shuaishuai Zhao, Yao Gao, Zhe Liu\\
 School of Software Engineering, Xi'an Jiaotong University, Xi'an, China\\
{\tt\small xiubai@xjtu.edu.cn; \{sszhaomh; yaogao; alfredliu\}@stu.xjtu.edu.cn} 
}

\maketitle

\begin{abstract}

Visual long-range interaction refers to modeling dependencies between distant feature points or blocks within an image, which can significantly enhance the model's robustness. Both CNN and Transformer can establish long-range interactions through layering and patch calculations. However, the underlying mechanism of long-range interaction in visual space remains unclear. We propose the mode-locking theory as the underlying mechanism, which constrains the phase and wavelength relationship between waves to achieve mode-locked interference waveform. We verify this theory through simulation experiments and demonstrate the mode-locking pattern in real-world scene models. Our proposed theory of long-range interaction provides a comprehensive understanding of the mechanism behind this phenomenon in artificial neural networks. This theory can inspire the integration of the mode-locking pattern into models to enhance their robustness.

\end{abstract}

\section{Introduction}

Visual long-range interaction refers to modeling dependencies of long-distance feature points or blocks within an image. The powerful visual long-range interaction capability can significantly improve the robustness of the model. Despite partial occlusion or interference, the model is able to accurately identify and locate specified objects by its long-range interaction capabilities to establish dependency relationships.

How does the existing neural network model establish the ability of visual long-range interaction? Table \ref{table1} shows various ways to model long-range interactions in artificial neural networks.
CNNs \cite{simonyan2014very, he2016deep} expand their receptive field through added layers to establish long-range dependencies in images, while RNNs establish dependencies through time series \cite{salinas2020deepar, rangapuram2018deep, wen2017multi}. In GNNs \cite{kipf2016semi, hamilton2017inductive, velivckovic2017graph, kipf2016variational}, distant nodes sequentially traverse paths on a graph to establish dependencies. In Transformers \cite{dosovitskiy2020image,liu2021swin, chen2021crossvit}, long-range patches directly establish dependencies based on patch distance. 
This method for interactions is exhaustive and crude, and cannot reflect the underlying mechanism of remote interaction in the visual system. 
Despite the operational-level modeling of long-range interactions, the underlying mechanism of long-range interaction in visual space remains unclear.

 \begin{figure}[t]
	\begin{center}
		\includegraphics[width = 1 \linewidth]{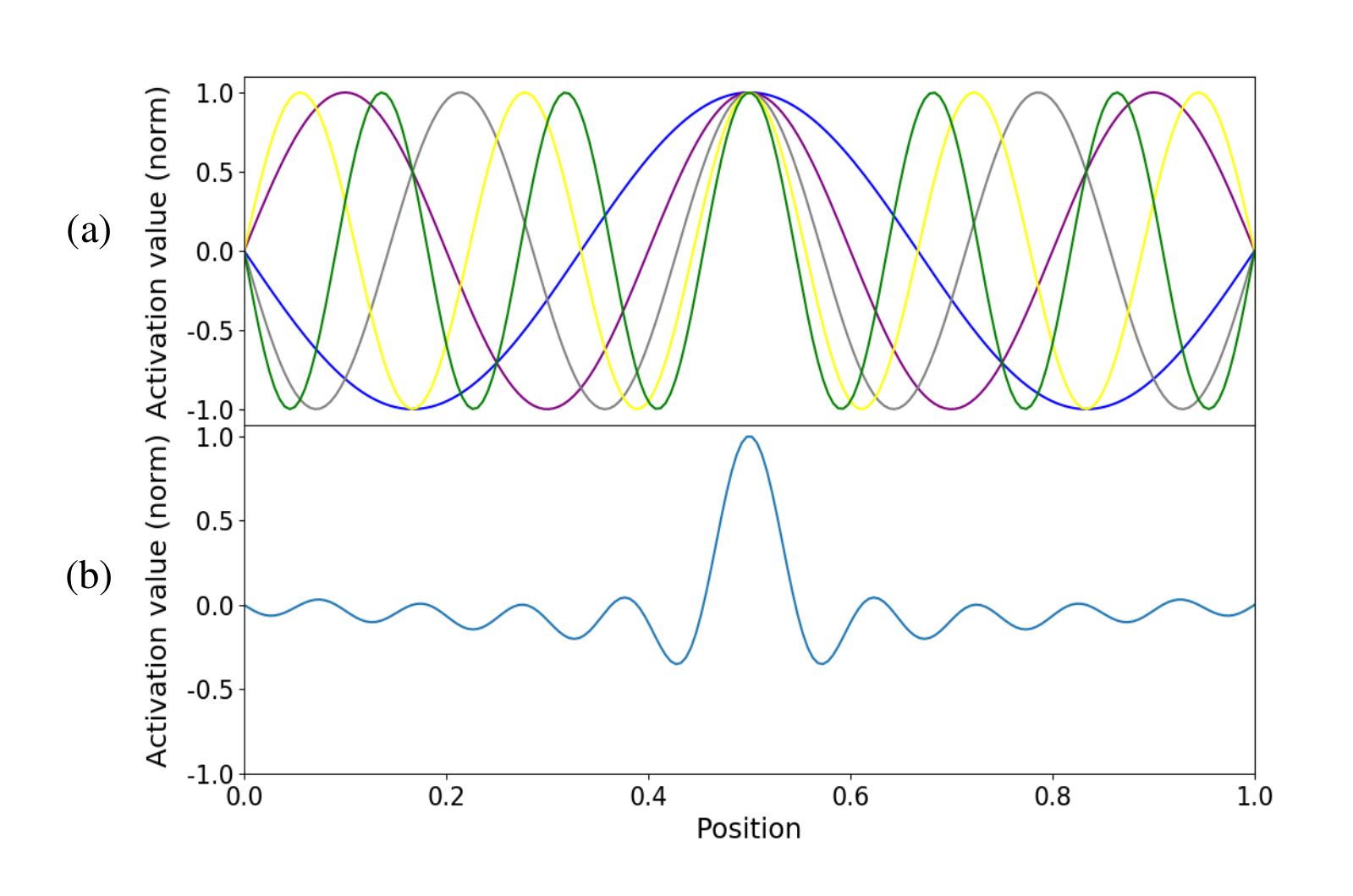}	
	\end{center}
	\caption{Diagram of the mode-locking theory for long-range interaction. (a) depicts a set of waves with a specific phase-wavelength relationship. (b) shows the waveform resulting from the mode-locking of waves in (a). When the small receptive field (a) reaches the state in (b), the large receptive field can perform visual long-range interaction. }
	\label{fig1}
\end{figure}

\begin{table*}[t]
	\begin{center}
		\begin{adjustbox}{width= 0.9 \linewidth}
        \scriptsize
		\begin{tabular}{llll}
        \toprule
        Types            & Inputs       & Relations      & Model long-range interactions  \\ \midrule
        CNN              & Grid elements        & Local          & Increase the number of layers  \\
        Transformer      & Patches     & Patch-to-patch     & Calculate the relationships between any two patches  \\
        RNN              & Timesteps     & Sequential     & Sequentially \\
        GNN              & Nodes     & Edges     & Calculate the path in the graph \\ \bottomrule
      \end{tabular}
	\end{adjustbox}
	\end{center}
 	\caption{Various ways to model long-range interactions in artificial neural networks}
		\label{table1}
\end{table*}

Experiments in \cite{Bai2020OnTR, bai2022emergence} verify that simple square waves significantly influence spatial-related models such as semantic segmentation, object detection, and skeleton detection. These models are sensitive to wave orientation and wavelength, and the introduction of attack samples containing specific waves affects model performance. Through this approach, researchers can identify the strongest attack wave that disrupts predictions or the weakest attack wave that improves predictions. 
Bai et al. \cite{bai2022emergence} discover that wave patterns can explain the high sensitivity of CNN models to wave properties in computer vision tasks involving spatial location. These findings suggest that wave patterns are inherent to CNN encoding of visual space.

Inspired by: 1) Wave patterns in spatial-related CNN models \cite{bai2022emergence};
2) Visual stereoscopic produced by two orthogonal phase waves in the brain's visual cortex \cite{ohzawa1990stereoscopic};
3) Mode-locking theory \cite{haus2000mode} based on wave interference in chirped pulse amplification technology (2018 Nobel Prize) in laser communication, introducing a fixed wavelength and phase relationship for waves to periodically build up interference and produce a series of pulses.

We first propose the mode-locking theory as the underlying mechanism for visual long-range interaction (Figure \ref{fig1}). This theory constrains the phase and wavelength relationships between waves to achieve a mode-locked interference waveform, thereby solving long-range interaction. Next, we verify this theory through a simulation experiment and demonstrate the mode-locking pattern in a real-world scene long-range interaction model.

The contributions of our paper are as follows.
\begin{itemize}
\item[$\bullet$] We first propose the mode-locking theory as the underlying mechanism for visual long-range interaction.
\item[$\bullet$] We verify this theory through a long-range interaction simulation experiment.
\item[$\bullet$] We discover the mode-locking pattern in a real-world scene long-range interaction model.
\end{itemize}

\begin{figure*}[t]
	\begin{center}
		\includegraphics[width = 0.85 \linewidth]{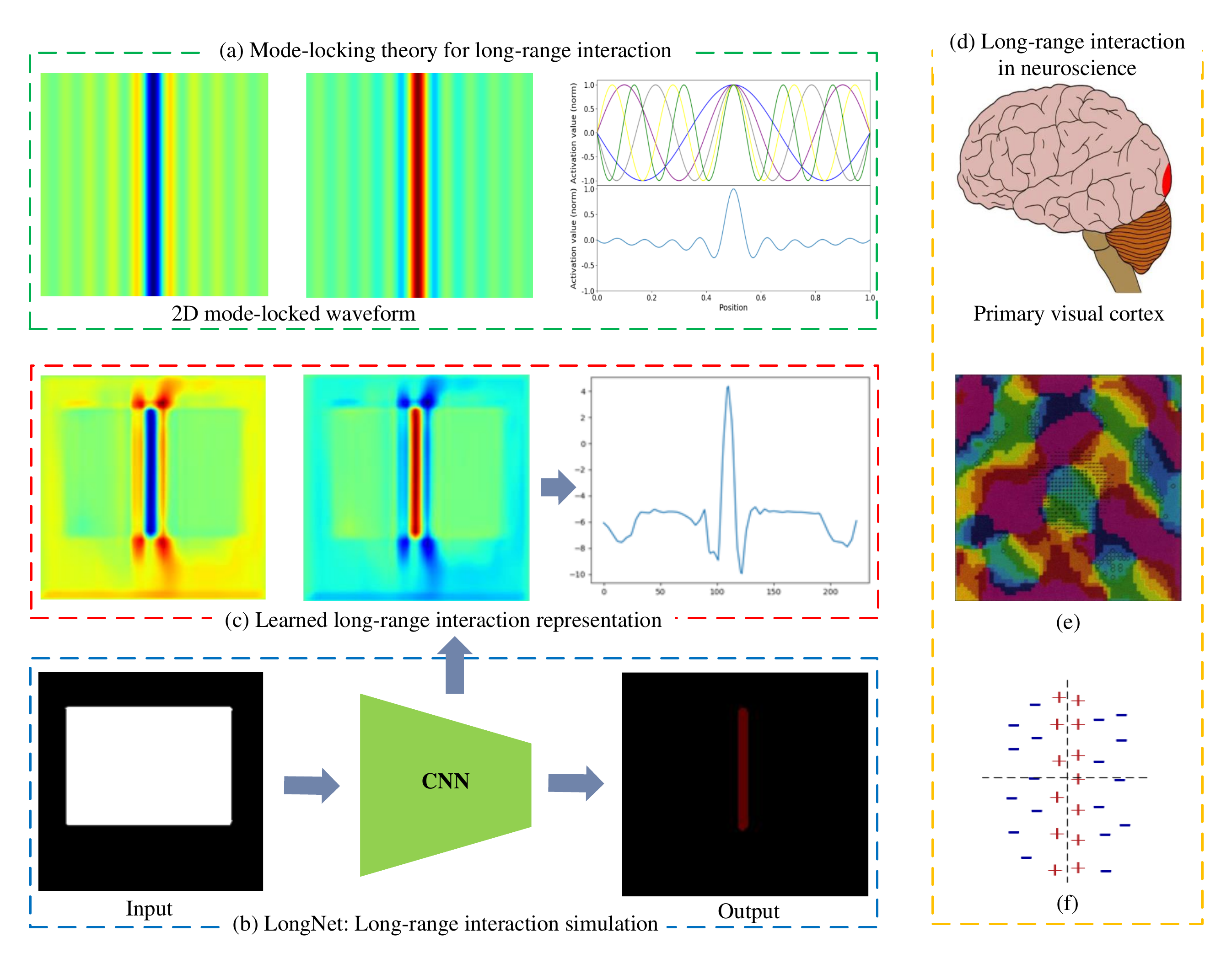}	
	\end{center}
	\caption{LongNet: a simulation experiment for visual long-range interaction. The task is to predict the spatial symmetry axis of a randomly generated input graph. The accuracy of this prediction indicates that the network has learned the long-range interaction relationship. The resulting activation pattern aligns with the mode-locking pattern and lateral interactions in the visual cortex. (e) Long-range lateral interaction map obtained by optical imaging in a macaque monkey \cite{stemmler1995lateral}. Each pixel's color corresponds to the preferred orientation of the respective cell (yellow, horizontal; green, $45^{\circ}$; red, $90^{\circ}$; and blue, $135^{\circ}$). Excitatory connections (+); inhibitory connections (-). Long-range connections are established from excitatory cells to both inhibitory and excitatory cells. (f) An illustration of a single orientation is chosen from (e). It includes intermediate excitation and lateral inhibition.}
	\label{fig2}
\end{figure*}

\section{Related work}

We first provide a brief overview of the impact of CNNs and Transformers on long-range interaction in visual neural network models.

\textbf{Convolutional Neural Networks.} Currently, many CNNs enhance their ability to establish effective long-range dependencies by increasing network depth. This is because deeper networks possess larger receptive fields, more abstract features, and better long-range dependency establishment. In the earlier methods \cite{krizhevsky2012imagenet, szegedy2015going, simonyan2014very}, the depth of the network is expanded by concatenating convolutional and pooling layers to expand the receptive field. However, this approach encountered limitations such as gradient vanishing and restricting network depth. To overcome this problem, He et al.\cite{he2016deep} introduce ResNet, which utilizes residual connections to enable networks to reach thousands of layers in depth. Meanwhile, Huang et al. \cite{huang2017densely} propose DenseNet, which introduces dense connections to facilitate feature reuse. Wang et al. \cite{wang2018non} propose non-local to avoid problems of continuously stacking network layers, allowing the response at each pixel to be the sum of feature weights across all points, improving long-distance dependencies.

\textbf{Vision Transformer.} While the Transformer operates globally, its shallow layers still tend to focus more on local features \cite{wang2021pyramid}. As the number of layers increases, the Transformer gradually shifts its focus toward capturing more global features. Swin Transformer \cite{liu2021swin} adopts a layered feature pooling strategy, allowing the shallow layers to extract local features while the deeper layers capture more abstract and global features. CrossViT \cite{chen2021crossvit} introduces an MLP module to enable cross-connections between low-level and high-level features. Additionally, the performance impact of deep remote interaction capability is evident in the PVT network \cite{wang2021pyramid}. Removing the self-attention mechanism from the first two layers only leads to a marginal drop in the Top accuracy rate on ImageNet from 80.9\% to 80.4\%. However, when the self-attention mechanism is removed from the subsequent two layers, the performance significantly drops to 66.8\%, highlighting the importance of deep long-range interaction capability for model performance \cite{wang2021pyramid}.

CNNs sequentially compute feature relationships, limiting globality and robustness due to their local operation. Transformers use self-attention mechanisms to capture feature correlations globally across all patches, resulting in good robustness but low efficiency, high computational complexity, a large amount of data required, and loss of local spatial information.

Most existing methods for modeling long-range interactions in artificial neural networks are limited to operational-level approaches. However, the underlying mechanisms of this neural processing are still unclear, and the principles of long-range interaction in artificial neural networks remain a mystery.

\section{Methods}

This section covers the mode-locking theory as the underlying mechanism for visual long-range interaction (Figure \ref{fig1}), its verification through a simulation experiment, and the demonstration of the mode-locking pattern in a real scene model.

\subsection{Mode-locking theory for long-range interaction}

Mode-locking theory \cite{haus2000mode} based on wave interference in the chirped pulse amplification technology in laser communication (2018 Nobel Prize). This theory refers to introducing a fixed wavelength, phase relationship, these waves periodically build up interference and produce a series of pulses.
Visual stereoscopic in the brain's visual cortex can be produced by two orthogonal phase waves \cite{ohzawa1990stereoscopic}, while wave patterns emerge in spatial-related CNN models \cite{bai2022emergence}.

Based on these findings, we propose the mode-locking theory as the underlying mechanism of visual long-range interaction (Figure \ref{fig1}), which utilizes phase and wavelength relationships between waves to achieve mode-locked interference waveforms, thus solving the problem of long-range interaction.
The specific principle of mode-locking is as follows.

\begin{figure*}[t]
	\begin{center}
		\includegraphics[width = 0.88 \linewidth]{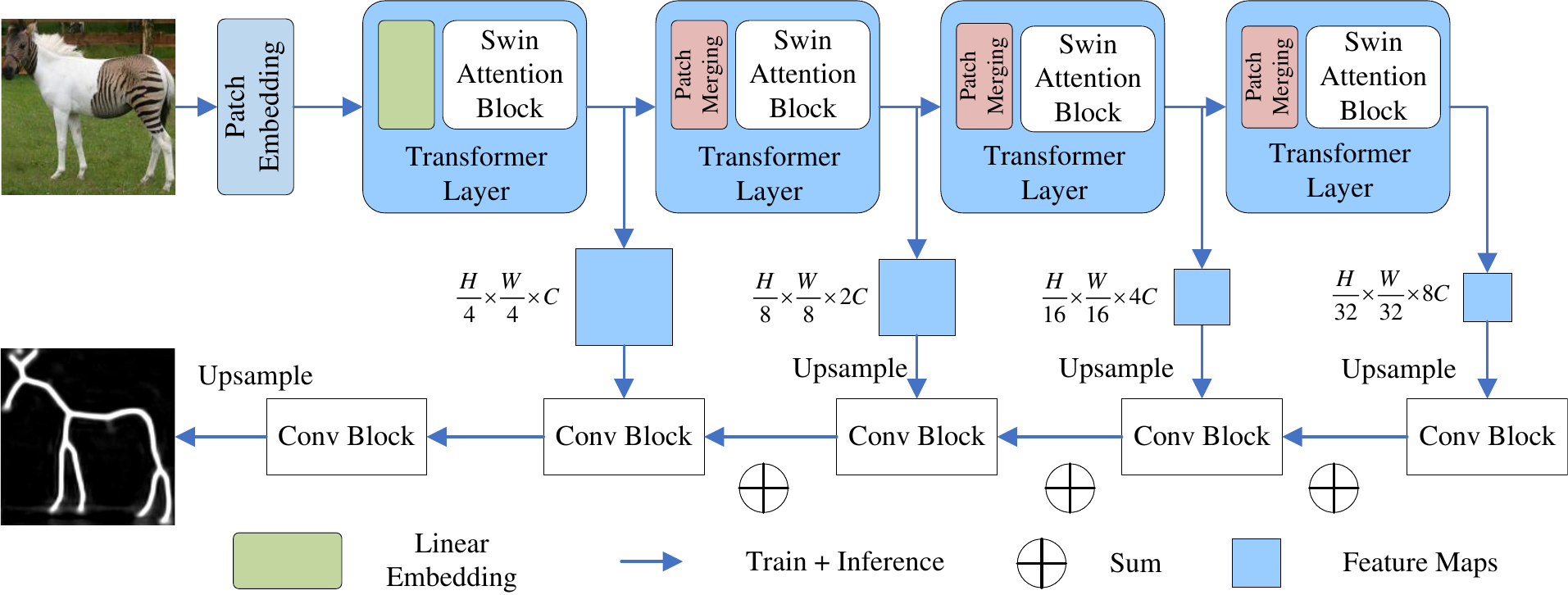}	
	\end{center}
	\caption{Architecture of our SkeSwin. The encoder consists of Swin Transformer, and the decoder consists of Sem-FPN.}
	\label{fig3}
\end{figure*}

A group of $sine$ waves in an image with a spatial distance of $L$ can be expressed as:
\begin{equation}y_i = A * \sin({\frac{2\pi}{\lambda_i}x + \varphi_i}), \quad i = 1,2,...,n \end{equation}
where $A$ denotes the amplitude, $\varphi_i$ denotes the phase, $\lambda_i$ denotes the wavelength, and $n$ denotes the number of waves.

When a group of waves with fixed wavelength and phase is at a spatial distance $L$ in an image, they can achieve mode-locking and periodically build up interference to produce a series of pulses. For mode-locking to occur, $L$ must be an odd multiple of half the wavelength of all waves, which can be expressed as:
\begin{equation}
    L = q \frac{\lambda}{2}, \quad \text{ $q$ is odd }
\end{equation}

To achieve mode-locking, the phase difference between two waves of adjacent wavelengths must be $\pi$. Therefore, the wave functions can be modified as follows:
\begin{equation}
    y_i = A * \sin({q \frac{1}{L} \pi x + i \% 2 * \pi}), \quad q = 2i + 1
\end{equation}
The mode-locked waves can be superimposed as $y_{sum}$:
\begin{equation}
    y_{sum} = \frac{1}{n} \sum_{i=1}^{n} y_i
\end{equation}

When small receptive fields $y_i$ (Figure \ref{fig1} (a)) reach the state in Figure \ref{fig1} (b), the large receptive field $y_{sum}$ can perform visual long-range interaction.
To capture large-scale objects in images, it can use a series of waveforms with fixed phase relationships and small receptive fields to create a mode-locked waveform, acting as a large receptive field filter that establishes correlations between both sides of the object.

\subsection{Long-range interaction simulation design (LongNet)}

Figure \ref{fig2} shows the long-range interaction simulation experimental design to verify the above mode-locking theory. The used CNN is called \textbf{LongNet}. Our central goal is to observe how long-distance feature pixels interact. To avoid the  perturbation of image textures and colors, We take as input a rectangular box of random size and location. Labels are the spatial geometric symmetry axes of the input graph.

If the convolutional neural network can predict the symmetry axis of the input graph, it can be simulated to indicate that the network has learned the long-range interaction relationship between two sides of the input graph. 
The spatial activation map learned by training a convolutional neural network is in good agreement with the theory of long-range interaction based on mode-locking, thus validating the above theory. 

Furthermore, the activation pattern observed in CNN is also consistent with the principle of lateral interaction in the visual cortex \cite{stemmler1995lateral}, which includes intermediate excitation and lateral inhibition.

After the (input, label) set is generated, it can force the LongNet to learn visual long-range interaction. 
 The network architecture of LongNet is based on DeepLabv3+ \cite{chen2018encoder}. The used backbone model is ResNet50 \cite{he2016deep}. 
Since this is an imbalanced segmentation problem, we use dice loss \cite{li2019dice} and cross-entropy loss.

\subsection{The actual task of long-range interaction (SkeSwin)}

Skeleton detection \cite{bai2019skeleton, BAI202311, xu2021deepflux} is a typical task that requires long-range interaction capabilities, that is, the skeleton feature contains the spatial scale information that depends strongly on the long-distance information in the image.

We focus on the skeleton detection task to observe long-range interaction patterns because other visual tasks, such as classification, semantic segmentation, and object detection, typically encode both space and object information, making it difficult to observe specific spatial interaction patterns. In contrast, skeleton detection output mainly encodes spatial structure information that heavily relies on remote interactions, making it easier to observe long-range interaction patterns in the model.

Transformers are known for their strong long-range interaction capabilities, as they can calculate the correlation between any two patches. We aim to observe whether there is a mode-locking pattern in the spatial activation pattern of a Transformer-based skeleton detection model.

To this end, we propose a model for skeleton detection, named \textbf{SkeSwin} (Figure \ref{fig3}), that combines the Swin Transformer network \cite{liu2021swin} and the Semantic Feature Pyramid Network (Sem-FPN) \cite{kirillov2019panoptic}.

Swin Transformer divides the encoder into four stages, using window self-attention and shift window methods to achieve global information interaction. Semantic FPN is adopted as the decoder for multi-scale feature fusion on the feature maps at each stage. The used loss is weighted $L_2$ loss.

\begin{figure*}[t]
	\begin{center}
		\includegraphics[width = 1 \linewidth]{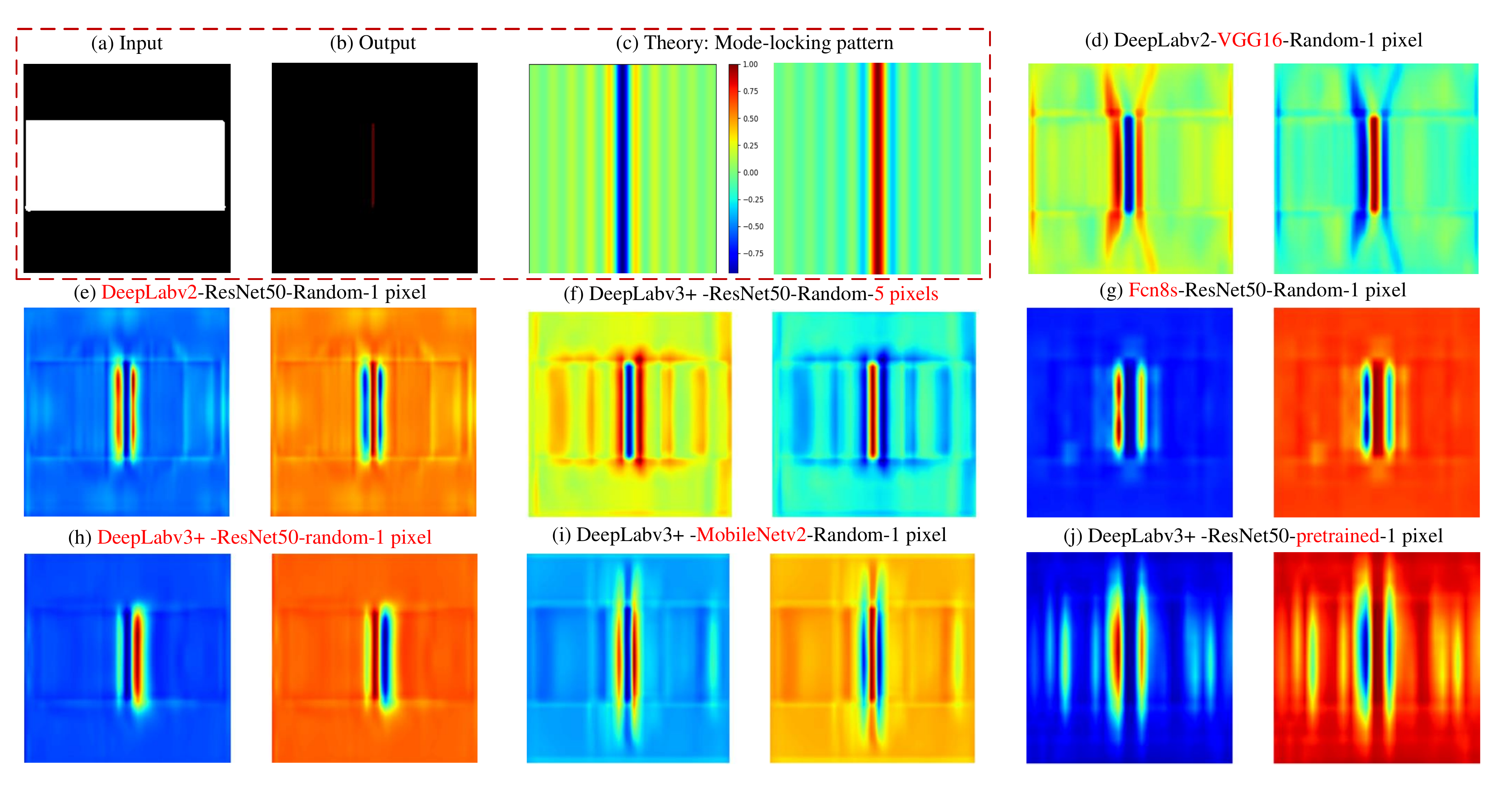}	
	\end{center}
	\caption{Mode-locking interference patterns emerge in networks trained to represent long-range interaction. (a) input. (b) output. (c) Theory：mode-locking pattern. (d)-(j): Feature maps in the last hidden layer exhibit spatial responses resembling mode-locking interference patterns.}
	\label{fig4}
\end{figure*}

\section{Experiments}

\subsection{Datasets}
\textbf{LongNet: }
For this simulation experiment, we randomly generate 10,000 rectangular images with varying sizes and positions as inputs, and use their spatial geometric symmetry axes as labels. Among these images, 8500 are randomly selected for training and 1500 are used for testing.

\textbf{SkeSwin:}
\textbf{SK-LARGE} \cite{shen2017deepskeleton}: It is sampled from the MS-COCO \cite{karpathy2015deep} dataset and contains 746 training samples and 745 testing samples.
\textbf{SYM-PASCAL} \cite{ke2017srn}: It is built on the semantic segmentation dataset PASCAL VOC 2011\cite{everingham2010pascal}, with 648 training samples and 787 samples.
\textbf{WH-SYMMAX} \cite{martin2001database}: It is stemmed from the Weizmann Horse dataset, which consists of all images of horses. There are 228 samples in the training set and 100 samples in the test set.
\textbf{SYMMAX300} \cite{tsogkas2012learning}: It is obtained from the BSDS300\cite{martin2001database} dataset, with 200 samples in the training set and 100 samples in the test set.

\subsection{Metric}
\textbf{LongNet: }
We utilize F-measure and mIoU as evaluation metrics for performance.

\textbf{SkeSwin:}
We adopt the F-measure to evaluate performance. F-measure is calculated by the formula: $F=2PR / (P+R)$, $P$: Precision, $R$: Recall. The results need to go through non-maximum suppression (NMS) \cite{dollar2014fast} to obtain the skeleton map, and use the thinning algorithm \cite{nemeth2011thinning} to obtain the final predicted skeleton.

 \begin{table*}[t]
\begin{center}
\renewcommand\arraystretch{2}
\begin{adjustbox}{width= 0.8 \linewidth}
\scriptsize
\begin{tabular}{l|l|l|c|c|c}
\toprule
Fixed components & Factors & Types & F-measure $\uparrow$ & mIoU $\uparrow$ & {Mode-locking pattern} \\ 
\hline
\multirow {2}{*}{\makecell[l]{-DeepLabv3+ \\ -ResNet50 \\ -1 pixel}} & \multirow {2}{*}{\makecell[l]{Initialization}} & Random & 0.560 & 0.385 & $\surd$ \\
\cline{3-6}
& &{Pretrained ImageNet}  & 0.633 & 0.457 & $\surd$\\
\cline{1-6}
\hline
\multirow {2}{*}{\makecell[l]{-DeepLabv3+ \\ -ResNet50 \\ -Random}} & \multirow {2}{*}{\makecell[l]{Label width}} & {1 pixel} & 0.560 & 0.385 & $\surd$ \\
\cline{3-6}
& &{5 pixels} & 0.869 & 0.768 & $\surd$ \\
\cline{1-6}
\multirow{3}{*}{\makecell[l]{-ResNet50 \\ -Random \\ -1 pixel}} &
\multirow{3}{*}{\makecell[l]{Architecture}} & DeepLabv3+ & 0.560 & 0.385 & $\surd$\\
\cline{3-6}
& & DeepLabv2 &  0.455 & 0.292 & $\surd$\\
\cline{3-6}
& & FCN8s &  0.501 & 0.334 & $\surd$\\
\cline{1-6}
\multirow{3}{*}{\makecell[l]{-DeepLabv3+ \\ -Random \\ -1 pixel}} & 
\multirow{3}{*}{\makecell[l]{Backbone }} & MobileNetv2 & 0.570 & 0.397 & $\surd$\\
\cline{3-6}
& & ResNet50& 0.560 & 0.385 & $\surd$\\
\cline{3-6}
& & VGG16& 0.519 & 0.333 & $\surd$\\
\bottomrule
\end{tabular}
\end{adjustbox}
\end{center}
\caption{Effect of various factors on the emergence of mode-locking pattern in LongNet. $\surd$ means it can observe mode-locking patterns.}
\label{table2}
\end{table*}

\begin{figure*}[t]
  \centering
  \includegraphics[width=1 \linewidth]{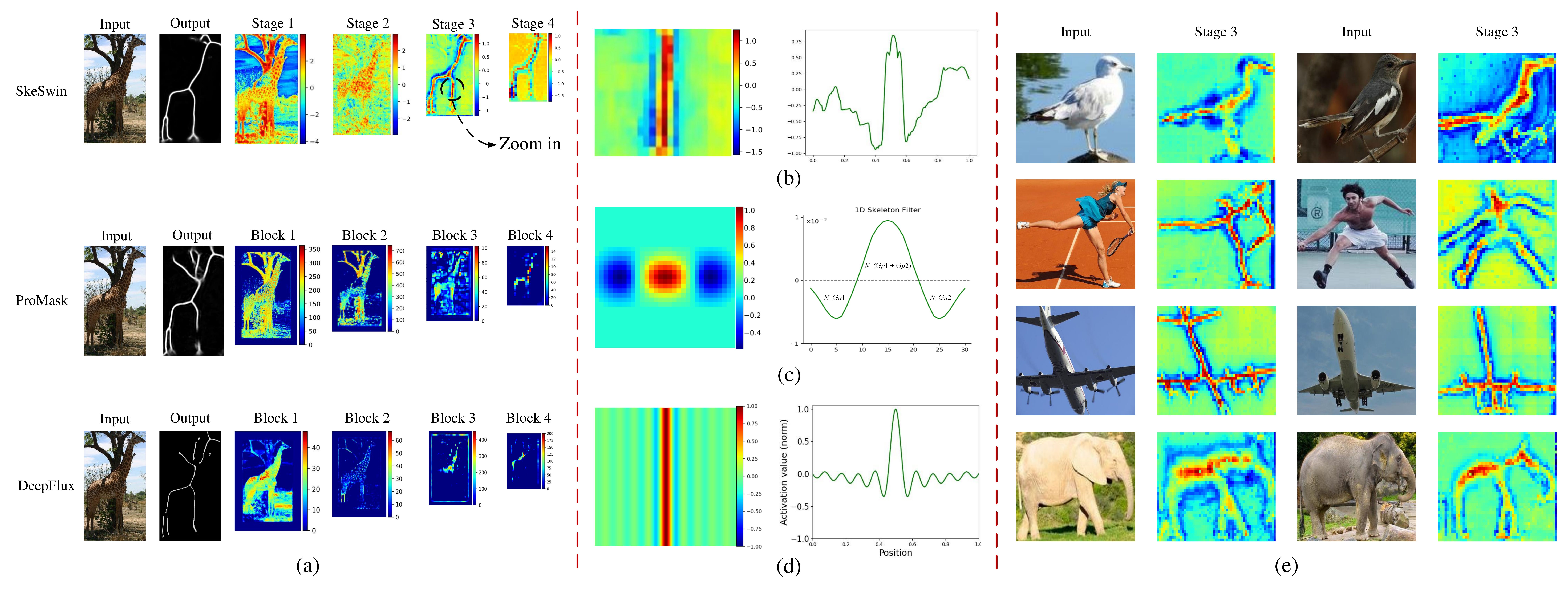}
  \caption{Visualization of feature maps. (a): Feature maps of SkeSwin (ours), ProMask \cite{BAI202311} and DeepFlux \cite{xu2021deepflux} at each stage. (b): 2D and 1D excitation regions of the skeleton, as well as the neighboring inhibition region. (c): 2D and 1D Skeleton Filter \cite{bai2019skeleton}. (d): 2D and 1D mode-locked waveforms. (e): More test samples of SkeSwin.}
  \label{fig5}
\end{figure*}

\begin{table*}[t]
	\begin{center}
		\begin{adjustbox}{width= 0.8 \linewidth}
        \scriptsize
		\begin{tabular}{lcccc}
        \toprule
        Method            & SK-LARGE       & WH-SYMMAX      & SYM-PASCAL     & SYMMAX300      \\ \midrule
        HED \cite{xie2015holistically}                             & 0.497     & 0.732     & 0.369     & 0.427 \\
        RCF \cite{liu2017richer}                             & 0.626     & 0.751     & 0.392     & -     \\
        FSDS \cite{shen2016object}                            & 0.633     & 0.769     & 0.418     & 0.467 \\
        SRN \cite{ke2017srn}                             & 0.678     & 0.780     & 0.443     & 0.446 \\
        OD-SRN \cite{liu2019orthogonal}                          & 0.676     & 0.804     & 0.444     & 0.489 \\
        LSN \cite{liu2018linear}                             & 0.668     & 0.797     & 0.425     & 0.480 \\
        Hi-Fi \cite{zhao2018hi}                           & 0.724     & 0.805     & 0.454     & 0.486 \\
        DeepFlux \cite{xu2021deepflux}   & 0.734         & 0.850     & 0.558     & 0.525 \\
        MSB+ \cite{yang2021msb}               & 0.754          & 0.864          & \textbf{0.587}       & 0.530          \\
        GeoSkeletonNet \cite{xu2019geometry}    & 0.757          & 0.849          & 0.520          & 0.501          \\
        AdaLSN \cite{liu2021adaptive}           & 0.786          & 0.851          & 0.497          & 0.495          \\ 
        ProMask \cite{BAI202311}           & 0.748          & 0.858          & 0.564          & 0.525          \\    \midrule
        SkeSwin (ours)  & \textbf{0.769}  & \textbf{0.900}  &  0.568      & \textbf{0.550}    \\ \bottomrule
      \end{tabular}
	\end{adjustbox}
	\end{center}
 	\caption{F-measure comparison between the Transformer method (SkeSwin) and other CNN-based skeleton detection methods. }
		\label{table3}
\end{table*}

\begin{figure*}[t]
  \centering
  \includegraphics[width=0.95 \linewidth]{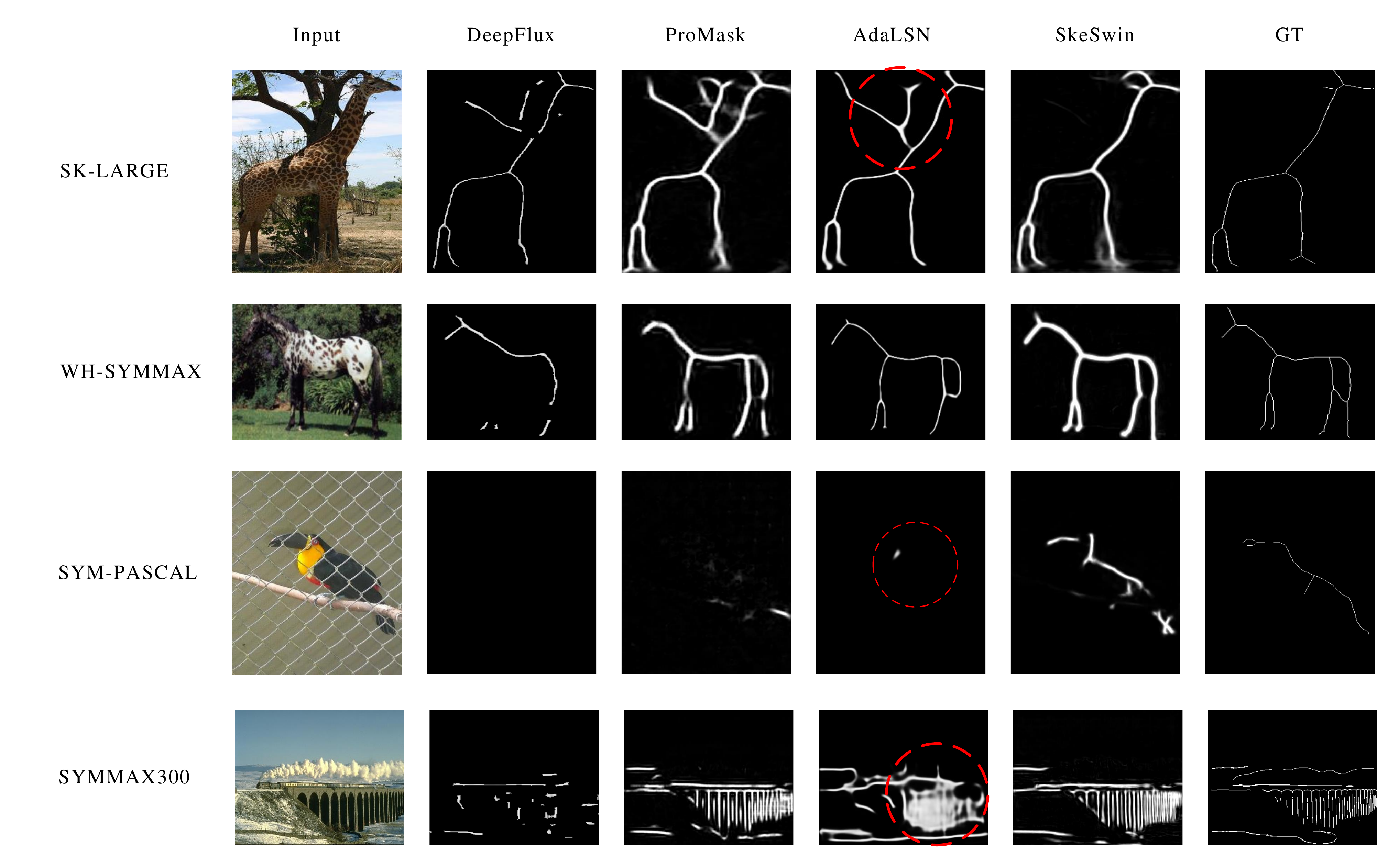}
  \caption{Qualitative comparison between Transformer method (SkeSwin) and CNN method (ProMask \cite{BAI202311}, DeepFlux \cite{xu2021deepflux}, AdaLSN \cite{liu2021adaptive}). Utilizing image long-range information in the detection model improves the model's robustness. For instance, even when a bird is partially occluded by a grid, SkeSwin can effectively detect the bird's skeleton. }
  \label{fig6}
\end{figure*}

\subsection{Implementation details}
\textbf{LongNet:}
Our model is trained using Adam optimizer with a step learning rate strategy initialized to 5e-4, while for deeplabv2 we use an initial learning rate of 5e-5. The batch size is 32, and we train for a total of 50 epochs. All training and testing are done using Pytorch and Nvidia RTX 2080Ti GPU.

\textbf{SkeSwin:}
SkeSwin is based on the open-source framework mmsegmentation \cite{chen2019mmdetection}.
We use offline and online data augmentation to adapt to the mini-batch training method of Transformer.
Offline data augmentation strategies are: 1) scale (0.8, 1.0, 1.2); 2) flip vertically and horizontally; 3) rotate in four directions (0°, 90°, 180°, 270°). 
The online data enhancement strategy is: 1) proportional scaling (0.5, 0.75, 1.0, 1.25, 1.5, 1.75); 2) cropping to 640x640 resolution. 

The batch size is set to 2, and the balance factor $\alpha=0.2$.
The number of iterations for SYMMAX300 is set to 80K and the rest to 160K. We use the AdamW \cite{loshchilov2017decoupled} optimizer with an initial learning rate of 1e-4, betas of (0.9, 0.999), weight decay of 1e-4, and a poly learning rate decay strategy.

\subsection{Analyses of network units: LongNet}

We have performed comprehensive experiments to confirm the presence of mode-locking patterns in CNN models trained to encode long-range interactions. As shown in Figure \ref{fig4}, interference patterns consistent with mode-locking theory emerge in the last hidden layer of the trained networks, as evidenced by spatially correlated responses. Specifically, panel (a) shows the input, panel (b) displays the output,  panel (c) shows the mode-locking pattern, while panels (d)-(j) illustrate the emergence of mode-locking patterns in feature maps of the last hidden layer.

The used configurations are as follows:

1) DeepLabv3+ \cite{chen2018encoder} -ResNet50 \cite{he2016deep} -Random-1 pixel

2) DeepLabv3+ -ResNet50-Pretrained-1 pixel

3) DeepLabv3+ -ResNet50-Random-5 pixels  

4) DeepLabv2 -ResNet50-Random-1 pixel  

5) Fcn8s \cite{long2015fully}-ResNet50-Random-1 pixel  

6) DeepLabv3+ -MobileNetv2 \cite{sandler2018mobilenetv2}-Random-1 pixel

7) DeepLabv2 \cite{chen2017deeplab}-VGG16 \cite{simonyan2014very}-Random-1 pixel

Table \ref{table2} presents the effects of various factors on the occurrence of mode-locking patterns in our LongNet.
The mode-locking pattern manifests itself in various network architectures, backbones, initialization methods, and label settings.

\subsection{Analyses of network units: SkeSwin}

Figure \ref{fig5} illustrates feature maps of SkeSwin, ProMask, and DeepFlux at each stage. The high-level feature maps of SkeSwin contain more semantic information related to the skeleton than ProMask and DeepFlux. In stage 3 or 4 of the feature maps in SkeSwin, the area adjacent to the skeleton has the lowest value, the background area has a middle value, and the skeleton excitation area has the highest value. This suggests that the network activates the skeleton regions and inhibits their neighborhoods. In contrast, the CNN model suppresses the background regions without making further judgments about the skeleton neighborhoods.

The feature map of SkeSwin in Figure \ref{fig5} (b) bears a striking resemblance to the waveform resulting from mode-locking (Figure \ref{fig5} (d)), further supporting the mode-locking theory as the underlying mechanism of visual long-range interaction.

In addition, the feature map of SkeSwin (Figure \ref{fig5} (b)) exhibits a strong resemblance to the waveform of Skeleton Filter (Figure \ref{fig5} (c)). This observation raises the question: why does the Transformer-based model share similarities with the human cortex's perception of skeleton detection?
There are several reasons:
1) There is evidence of skeleton signals in the V1 and IT visual cortices at the neural level \cite{hung2012medial}.
2) Orientation features in the V1 visual cortex are encoded using a similar pattern of Gabor filters \cite{jones1987evaluation}.
3) The Skeleton Filter is composed of Gabor-like filters in opposite directions.
4) The presence of Gabor filters in the human visual system and the simple architecture of the Skeleton Filter can explain the strong capabilities of humans in perceiving object skeletons, even under noisy conditions.
5) Therefore, the Skeleton Filter \cite{bai2019skeleton} may potentially be the configuration in the human visual system for recognizing skeleton features.
Therefore, the Transformer-based model produces a representation mechanism that is similar to the human brain in encoding skeleton features, which makes Transformer's approach more effective in dealing with complex samples.

Figure \ref{fig6} and Table \ref{table3} show the qualitative and quantitative comparison between Transformer-based skeleton detection and CNN-based skeleton detection methods, respectively.
Experimental results demonstrate that SkeSwin improves the performance of skeleton detection, outperforming state-of-the-art CNN models on commonly used datasets. 
Incorporating long-range image information in the detection model enhances its robustness.
For instance, even when a bird is partially occluded by a grid, SkeSwin can effectively detect the bird's skeleton.

\subsection{Discussion}

The proposed mode-locked pattern aligns with the principle of lateral interaction in the visual cortex \cite{stemmler1995lateral}, which involves intermediate excitation and lateral inhibition.
The lateral inhibition enhances object separation from their visual backgrounds, especially for object boundaries and outlines, improving visual sensitivity and contrast differences. It also confers strong robustness by forming an inhibition zone around highly activated pathways, preventing disturbances in a set of parallel neural pathways.

In artificial neural networks, the intermediate excitation with lateral inhibition (mode-locked pattern) offers advantages over the pattern with only intermediate excitation, including:
1) Enhanced spatial positioning: The lateral inhibition mechanism enables a more accurate representation of the object's spatial position, enhancing the model's spatial perception.
2) Improved robustness: Lateral inhibition stabilizes the intermediate excitation, suppressing perturbations and improving the model's robustness.

To achieve long-range dependencies, the Transformer-based skeleton detection model calculates the relationship between any two patches, resulting in lateral interaction in the feature map. 
However, the forming process in Transformer is a black box, and the underlying mechanism is still unclear.
Our proposed mode-locking theory provides an explanation for the lateral interaction phenomenon.

Unlike the Transformer, the CNN-based skeleton detection models for real scenes do not exhibit this mode-locked pattern. We believe this is due to the weak remote interaction ability of CNNs and the perturbation of complex semantic information in natural scenes. 
Therefore, we design a simple long-distance interactive simulation task to eliminate semantic perturbations, and successfully observe the locking mode in CNNs, thus verifying its validity.

\textbf{Potential value:} 
Our research inspires follow-up researchers to use the mode-locking theory to improve the long-range interaction ability of the CNN model, leading to more powerful global modeling and strong robustness with efficient calculation.
Tasks like object detection, pose estimation, and skeleton detection require strong spatial perception for accurately locating target points and building long-distance relationships. By incorporating mode-locking patterns, CNN models in these fields can be enhanced for improved performance and robustness.
Further exploration is warranted to uncover its potential practical applications.

\section{Conclusion}

The visual cortex possesses a crucial structural characteristic called lateral inhibition, which offers long-range interaction and improves robustness against perturbations. Artificial neural networks also exhibit this type of inhibition in their activation maps. We introduce the mode-locking theory as an explanation for the formation of this structure. Our proposed theory of long-range interaction can encourage the incorporation of the mode-locking principle into models aimed at enhancing robustness.

\section*{Acknowledgment}
This work was supported by the National Natural Science Foundation of China (No. 61802297) and the China Postdoctoral Science Foundation(No. 2021M702598).

{\small
\bibliographystyle{ieee_fullname}
\bibliography{scibib}
}

\end{CJK}
\end{document}